\documentclass[final,3p,times,12pt]{elsarticle} %
\usepackage{hyperref}
\usepackage{amsmath}
\usepackage{amsfonts}
\usepackage{graphicx}
\usepackage{algorithm}
\usepackage{algorithmicx}
\usepackage{algpseudocode}
\usepackage{booktabs}
\usepackage{array}
\usepackage{multirow}
\usepackage{xcolor}
\usepackage{colortbl}
\usepackage{titlesec}
\usepackage{epsfig}
\usepackage{epstopdf}
\usepackage{caption}
\usepackage{footmisc}
\usepackage{setspace}
\usepackage{txfonts}
\usepackage{lineno}
\usepackage{tikz}
\newcommand*{\circled}[1]{\lower.7ex\hbox{\tikz\draw (0pt, 0pt)%
    circle (.4em) node {\makebox[.3em][c]{\scriptsize #1}};}}

\begin{document}
\begin{frontmatter}
\title{Learning Reinforced Attentional Representation for End-to-End Visual Tracking}
\author[hitsz,hit]{Peng Gao}
\author[hitsz]{Qiquan Zhang}
\author[hitsz]{Fei Wang}
\author[hitsz,hit]{Liyi Xiao}
\author[vn,es,jp]{Hamido Fujita}
\author[hitsz]{Yan Zhang}
\address[hitsz]{School of Electronics and Information Engineering, Harbin Institute of Technology, Shenzhen, China}
\address[hit]{School of Astronautics, Harbin Institute of Technology, Harbin, China}
\address[vn]{Faculty of Information Technology, Ho Chi Minh City University of Technology (HUTECH), Ho Chi Minh City, Vietnam}
\address[es]{Andalusian Research Institute in Data Science and Computational Intelligence (DaSCI), University of Granada, Granada, Spain}
\address[jp]{Faculty of Software and Information Science, Iwate Prefectural University, Iwate, Japan}
\begin{abstract}
Although numerous recent tracking approaches have made tremendous advances in the last decade, achieving high-performance visual tracking remains a challenge. In this paper, we propose an end-to-end network model to learn reinforced attentional representation for accurate target object discrimination and localization. We utilize a novel hierarchical attentional module with long short-term memory and multi-layer perceptrons to leverage both inter- and intra-frame attention to effectively facilitate visual pattern emphasis. Moreover, we incorporate a contextual attentional correlation filter into the backbone network to make our model trainable in an end-to-end fashion. Our proposed approach not only takes full advantage of informative geometries and semantics but also updates correlation filters online without fine-tuning the backbone network to enable the adaptation of variations in the target object’s appearance. Extensive experiments conducted on several popular benchmark datasets demonstrate that our proposed approach is effective and computationally efficient.
\end{abstract}
\end{frontmatter}
\begin{spacing}{1.0}
\section{Introduction}\label{sec1}

Visual tracking is an essential and actively researched problem in the field of computer vision with various real-world applications such as robotic services, smart surveillance systems, autonomous driving, and human-computer interaction. It refers to the automatic estimation of the trajectory of an arbitrary target object, usually specified by a bounding box in the first frame, as it moves around in subsequent video frames. Although considerable progress has been made in last decade~\cite{survey2014,survey2018}, visual tracking is still commonly recognized as a very challenging task, partially due to numerous complicated real-world scenarios such as scale variations, fast motion, occlusions, and deformations.

One of the most successful tracking frameworks is the discriminative correlation filter (DCF)~\cite{kcf,dsst,cact}. With the benefits of fast Fourier transform, most DCF-based approaches can employ large numbers of cyclically shifted samples for training, and achieve high accuracy while running at impressive frame rates. Recent years have witnessed significant advances in convolutional neural network (CNN) on many computer vision tasks such as image classification and object detection~\cite{imagenet}. This is because the CNN can gradually proceed from learning finer-level geometries to coarse-level semantics of the target objects by transforming and enlarging the receptive fields at different convolutional layers~\cite{dlnn}. Encouraged by these great successes, some DCF-based trackers resort to using pre-trained CNN models~\cite{alexnet,vgg,resnet} instead of conventional handcrafted features~\cite{csk,kcf} for target object representation, and achieved favorable performance~\cite{csot,eco}. Recently, record-breaking performance and efficiency has been achieved using Siamese matching networks~\cite{siamfc,cfnet,siamrpn} for visual tracking. In each frame, these trackers learn a similarity metric between the target template and the candidate patch from current searching frame in an end-to-end fashion.
\begin{figure}[t]
\begin{center}
\includegraphics[width=\linewidth]{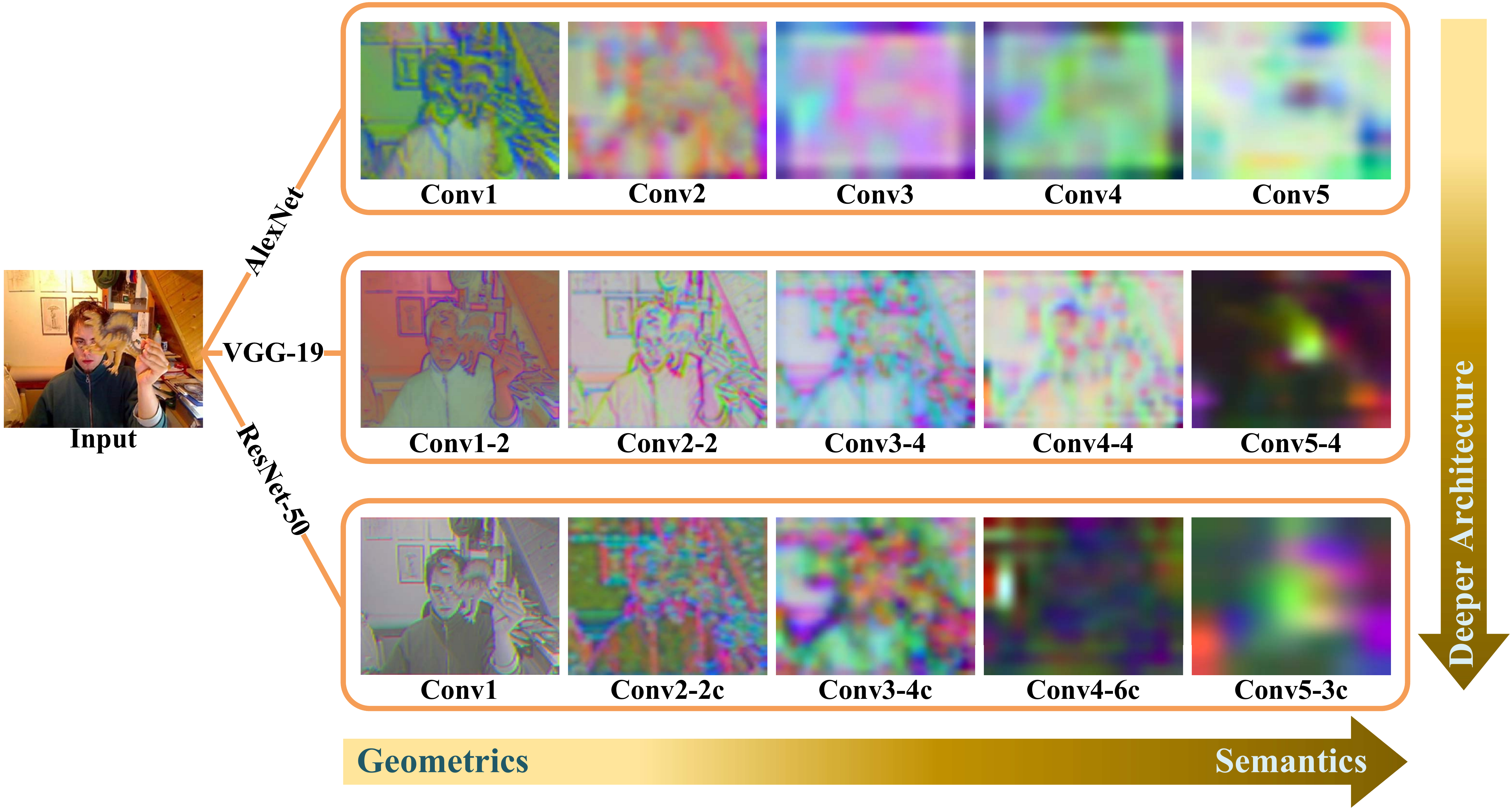}
\end{center}
\caption{Visualization of deep feature maps from different convolutional layers of different CNN architectures, including AlexNet~\cite{alexnet} (top row), VGG-19~\cite{vgg} (middle row) and ResNet-50~\cite{resnet} (bottom row). It is evident that low-level geometries from shallow layers, such as `\emph{conv}1' in AlexNet, `\emph{conv}1-2' in VGG-19 and `\emph{conv}1' in ResNet-50, remain fine-grained target-specific details, while high-level semantics from deep layers, such as `\emph{conv}5' in AlexNet, `\emph{conv}5-4' in VGG-19 and `\emph{conv}5-3c' in ResNet-50, contain coarse category-specific information. Compared with AlexNet, the architecture of ResNet-50 is deeper and more sophisticated. The example frame is shown from the sequence \emph{dinosaur}.}
\label{fig:feature}
\end{figure}
\begin{figure}[!t]
\begin{center}
\includegraphics[width=\linewidth]{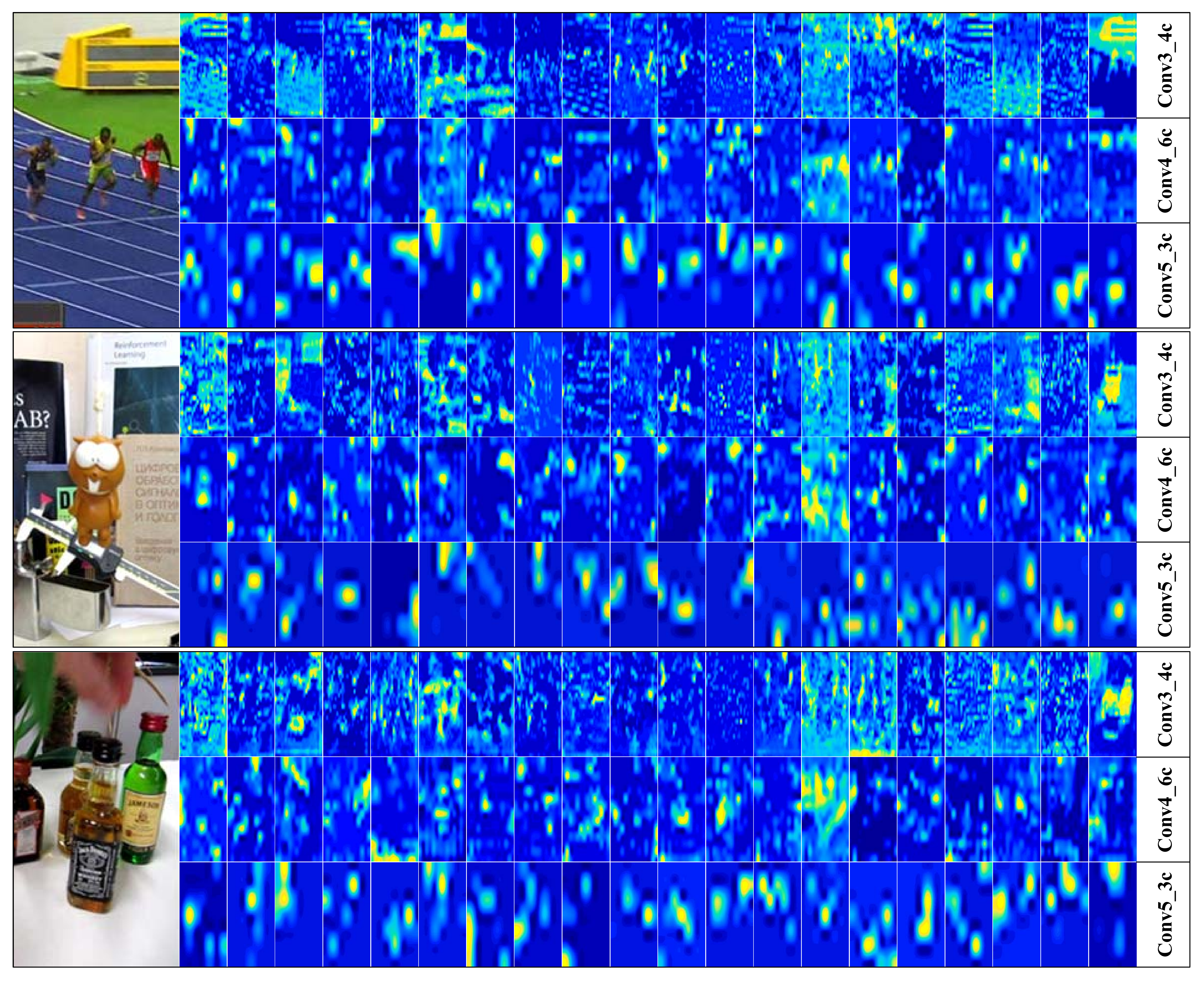}
\end{center}
\caption{Visualization of feature channels in the last layer of the `\emph{conv}3', `\emph{conv}4' and `\emph{conv}5' stages in ResNet-50~\cite{resnet}. Example frames are randomly picked up from \emph{Bolt}, \emph{Lemming} and \emph{Liquor} sequences (shown from top to bottom on the left). We show the features extracted from the 20 random channels of each stage from top to bottom on the right of the corresponding example frame. It is clear that only few of feature channels and regions contribute to target object representation, others may serve as information redundancy. A noteworthy is that for each example frame, the channels in the corresponding stage are the same.}
\label{fig:feat}
\end{figure}

Despite the significant progress mentioned above, existing CNN-based tracking approaches are still limited by several intractable obstacles. Most methods directly utilize off-the-shelf CNN models pre-trained on large-scale image classification datasets~\cite{imagenet,coco} to obtain generic representation of the target object~\cite{alexnet,vgg}. It is well acknowledged that different convolutional layers of CNNs, as shown in Fig.~\ref{fig:feature}, encode different types of features~\cite{dlnn}. Although features taken from the higher convolutional layers retain rich coarse-level category-specific semantics, they are ineffective for the accurately localizing or estimating the scale of the target object. Conversely, features extracted from the lower convolutional layers maintain more fine-level geometries to capture target-specific spatial details which facilitate accurately locating the target object, but are insufficient to distinguish objects from non-objects with similar characteristics. With the aim of best exploiting deep features, some prior works~\cite{hcft,hdtt,ccot,csot} have attempted to integrate advantages of fine-level geometries and coarse-level semantics using multiple refinement strategies. Unfortunately, compared with state-of-the-art approaches~\cite{siamfc,siamrpn} that only employ the outputs of the last layers to represent the target objects, their performance still has a notable gap. Combining features directly from multiple convolutional layers is thus not sufficient for representing target objects; they also tend to underperform under challenging scenarios.

Moreover, on deep feature maps, each feature channel corresponds to a particular type of visual pattern, whereas feature spatial regions represent object-specific details~\cite{song2018neural,zheng2019virtually}. We observe that deep features directly extracted from pre-trained CNN models treat every pixel equally along the channel-wise and spatial axes. Specifically, there is the possibility that only some of the features are closely related to the task of distinguishing specific target objects from background surroundings, others may be redundant information that may cause model drift, and probably lead to failures of tracking~\cite{mlcft,lcsn}, as illustrated in Fig.~\ref{fig:feat}. Recently, visual attention mechanism has brought remarkable progress to recent researches and performs surprisingly well in many computer vision tasks~\cite{senet,cbam}, owing to its ability to model contextual information. Although it is necessary to highlight useful features and suppress irrelevant information using attention mechanisms for visual tracking, some previous trackers~\cite{csrdcf,rasnet,flowtrack} only take advantage of intra-frame attention to learn which semantic attribute to select from the proper visual patterns along the channel axis, and do not take care about where to focus along the spatial axis, thus achieving inferior tracking results. Moreover, most existing CNN-based trackers implement their models with shallow networks such as AlexNet~\cite{alexnet}, they cannot exploit the benefits of more powerful representations from deeper networks such as ResNet~\cite{resnet}.
\begin{figure}[t]
\begin{center}
    \includegraphics[width=\linewidth]{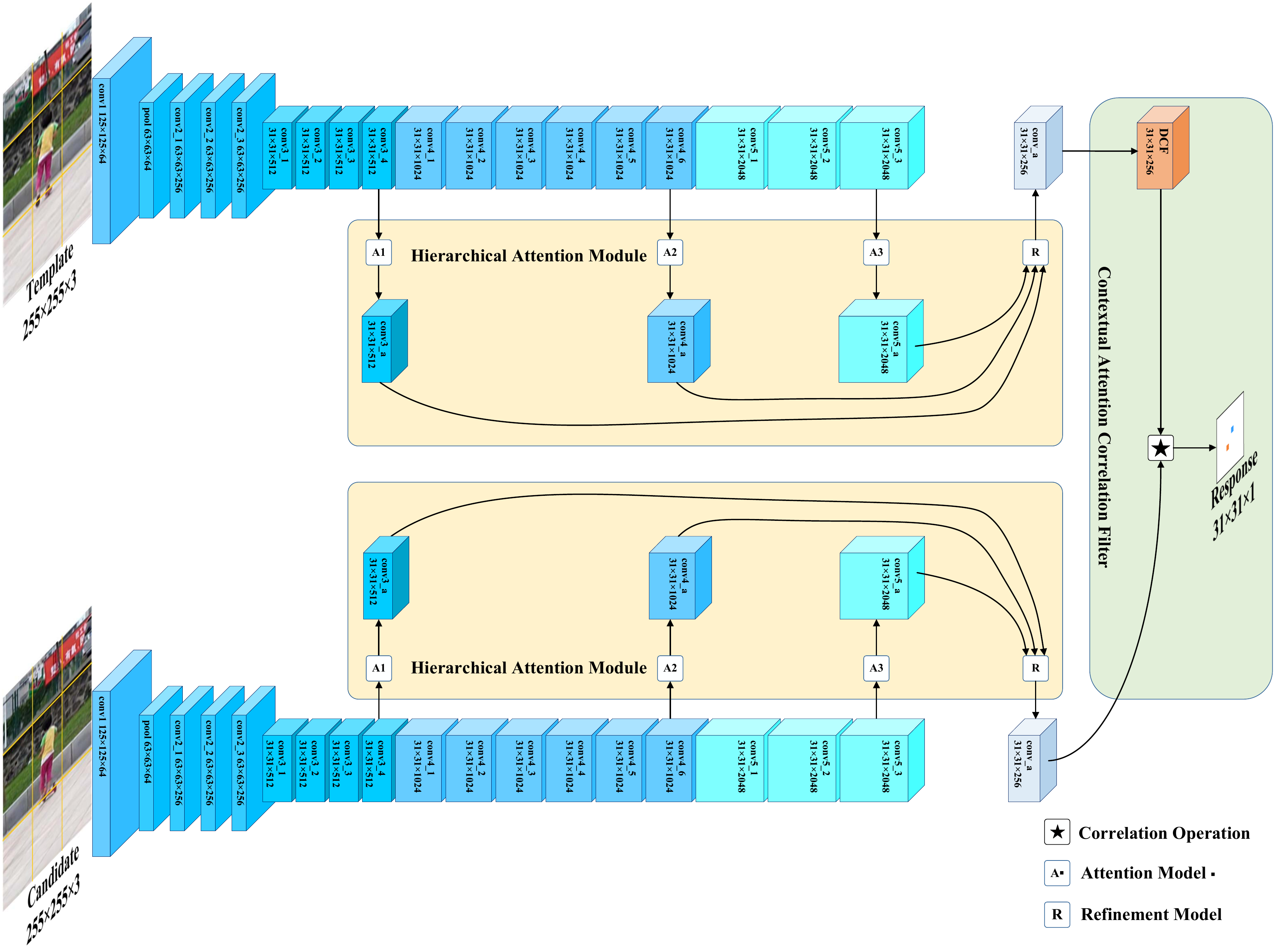}
\end{center}
\caption{The framework of the proposed tracking approach. Specifically, our approach contains three main components, i.e., a backbone network for deep feature extraction (detailed in Section~\ref{sec:3.1}), a hierarchical attention module for informative feature emphasis (detailed in Section~\ref{sec:3.2}), and a decision module for target object discrimination and localization (detailed in Section~\ref{sec:3.3}).}
\label{fig:arch}
\end{figure}

Notably, as the target objects could be anything, the pre-trained CNN models may be agnostic about some target objects not present in the training set. To ensure high-performance visual tracking, most trackers only employ the original deep features taken from the first frame to match candidate patches in subsequent frames~\cite{siamfc,siamrpn}. The characteristics of the target object are consistent within consecutive frames, and there exists a strong temporal relationship between the target object appearance and motion in video data~\cite{review1,review2}. Using contexts from historical frames may enhance tracking accuracy and robustness under challenging scenarios such as occlusions and deformations. The recurrent neural network (RNN), especially long short-term memory (LSTM)~\cite{lstm}, has achieved great success in many natural language processing (NLP) applications by saving attractive temporal cues and discarding irrelevant ones using prejudiced memory components, and thereby becoming suitable for exploring inter-frame attention during visual tracking. However, there are limited approaches that employ such network models in visual tracking~\cite{mam}. Most trackers ignore the inter-frame attention, and can hardly obtain appearance variations of the target objects well, which may lead to model drift. On the whole, how to take full use of inter- and intra-frame attention for visual tracking is a largely underexplored domain.

To address the above issues, we propose a unified end-to-end reinforced attentional Siamese network model, dubbed RAR, to pursue high-performance visual tracking. The framework of the proposed approach is shown in Fig.~\ref{fig:arch}. As abovementioned, it has already been proven that tracking can benefit from leveraging deep feature hierarchies across multiple convolutional layers~\cite{hcft,hdtt}. Therefore, we use a carefully modified ResNet-50 as the backbone network, and take multi-level deep features from the last three convolutional stages to enhance the effectiveness of target object representation. We adopt the tracking-by-detection paradigm to trace target objects, and reformulate the tracking problem as a sequential inference task. To emphasize informative representation and suppress information redundancy, we design a hierarchical attention module for learning multiple visual attention, which is composed of an inter-frame attention model and an intra-frame attention model. The inter-frame attention model is built upon convolutional LSTM units that can fully explore the temporal cues of the target object’s appearance at different convolutional layers in consecutive frames~\cite{mam,rtt}. It can be decomposed into sequential blocks, each of them corresponding to a specific time slice. We then design an intra-frame attention model that consists of two multi-layer perceptrons (MLPs) along the channel-wise and spatial axes on the deep feature maps~\cite{satin,cbam}. Through the inter- and intra-frame attention, we can obtain significantly more powerful attentional representations. It is worth noting that both the inter- and intra-frame attention are obtained separately at different convolutional layers. Subsequently, the hierarchical attentional representations are merged to produce a refined one using a refinement model made of convolutional layers and element-wise additions, rather than exploiting them independently or combining them directly. Specifically, the refined output is generated by successively integrating attentional representations from the last layer with that from earlier layers in a stacked manner. With the refinement model, we can obtain stronger representations that maintain coherent target-specific geometries and semantics at a desirable resolution. Besides, we adopt the DCF to discriminate and locate target objects. Because the background context around target objects has a significant impact on tracking performance, a contextual attentional DCF is employed as the decision module to take global context into account, and further eliminate unnecessary disturbance. To allow the whole network model to be trained from end to end, the correlation operation is reformulated as a differentiable correlational layer~\cite{cfnet,edcf}. Thus, the contextual attentional DCF can be updated online without fine-tuning the network model to guide the adaptation of the target object’s appearance model.

We summarize the main contributions of our work as follows:
\begin{enumerate}
    \item An end-to-end reinforced attentional Siamese network model is proposed for high-performance visual tracking.
    \item A hierarchical attention module is utilized to leverage both inter- and intra-frame attention at each convolutional layer to effectively highlight informative representations and suppress redundancy.
    \item A contextual attentional correlation layer that can take global context into account and further emphasize interesting regions is incorporated into the backbone network.
    \item Extensive and ablative experiments on four popular benchmark datasets, i.e., OTB-2013~\cite{otb2013}, OTB-2015~\cite{otb2015}, VOT-2016~\cite{vot2016} and VOT-2017~\cite{vot2017}, demonstrate that our proposed tracker outperforms state-of-the-art approaches.
\end{enumerate}

The rest of the paper is organized as follows. Section~\ref{sec:2} briefly reviews related works. Section~\ref{sec:3} illustrates the proposed tracking approach. Section~\ref{sec:4} details experiments and discusses results. Section~\ref{sec:5} concludes the paper.

\section{Related works}\label{sec:2}

Many real-world applications require visual tracking approaches with excellent effectiveness and efficiency. In this section, we briefly review tracking-by-detection methods based on the DCF and CNN, which are most related to our work. For other visual tracking methods, please refer to more comprehensive reviews~\cite{survey2014,survey2018}.

In the past few years, some tracking approaches that train DCFs by exploiting the properties of circular correlation and performing operations in the Fourier frequency domain has played a dominant role in the visual tracking community, because of their superior computational efficiency and reasonably good accuracy. Several extensions have been proposed to considerably improve tracking performance using multi-dimensional features~\cite{csk}, nonlinear kernel correlation~\cite{kcf}, robust scale estimation~\cite{dsst} and by reducing the boundary effects~\cite{cacf}. However, earlier DCF-based trackers take advantage of conventional handcrafted features~\cite{csk,kcf}, and thus suffer from inadequate representation capability.

Recently, with the rapid progress in deep learning techniques, CNN-based trackers have achieved remarkable progress, and become a trend in visual tracking. Some approaches incorporate CNN features into the DCF framework for tracking, and demonstrate outstanding accuracy and high efficiency. As previously known, the finer-level features that detail the spatial information play a vital role in accurate localization, and the coarse-level features that characterize semantics play a pivotal role in robust discrimination. Therefore, it is necessary to design a specific feature refinement scheme before discrimination. HCF~\cite{hcft} extracts the deep features from the hierarchical convolutional layers, and merges those features using a fixed weight scheme. HDT~\cite{hdtt} employs an adaptive weight to combine the deep features from multiple layers. However, these trackers merely exploit the CNN for feature extraction, and then learn the filters separately to locate the target object. Therefore, their performance may be suboptimal. Some later works attempt to train a network model to perform both feature extraction and target object localization simultaneously. Both CFNet~\cite{cfnet} and EDCF~\cite{edcf} unify the DCF as a differentiable correlation layer in a Siamese network model~\cite{siamfc}, and thus make it possible to learn powerful representation from end to end. These approaches have promoted the development of visual tracking, and greatly improved tracking performance. Nevertheless, many deep features taken from pre-trained CNN models are irrelative to the task of distinguishing the target object from the background. These disturbances will significantly limit the performance of the abovementioned end-to-end tracking approaches.

Instead of exploiting deep vanilla features for visual tracking, methods using attention weighted deep features alleviate model drift problems caused by background noise. In fact, when tracking a target object, the tracker should merely focus on a much smaller subset of deep features which can effectively distinguish and locate the target object from the background. This implies that many deep features are irrelative to representing the target object. Some works explore attention mechanisms to highlight useful information in visual tracking. CSRDCF~\cite{csrdcf} constructs a unique spatial reliability map to constrain filters learning. ACFN~\cite{acfn} establishes a unique attention mechanism to choose useful filters during tracking. RASNet~\cite{rasnet} and FlowTrack~\cite{flowtrack} further introduce an attention network similar to the architecture of SENet~\cite{senet} to enhance the representation capabilities of output features. Specifically, FlowTrack also clusters motion information to exploit historical cues. CCOT~\cite{ccot} takes previous frames into account during filter training to enhance its robustness. RTT~\cite{rtt} learns recurrent filters through an LSTM network to maintain the target object’s appearance. Nonetheless, all these trackers take advantage of only one or two aspects of attention to refine deep output features, exceedingly useful information in intermediate convolutional layers has not yet been fully explored.

Motivated by the above observations, we aim to achieve high-performance visual tracking by learning efficient representation and DCF mutually in an end-to-end network. Our approach is related to but different from EDCF~\cite{edcf} and HCF~\cite{hcft}. The former proposes a fully convolutional encoder-decoder network model to jointly perform similarity measurement and correlation operation on multi-level reinforced representation for multi-task tracking, but our approach additionally learn both inter- and intra-frame attention based on convolutional LSTM units and MLPs to emphasize useful features, and take global context and temporal correlation into account to train and update the DCF. The latter utilizes hierarchical convolutional features for robust tracking. However, rather than using a fixed weight scheme to fuse features from different levels, we first perform attentional analysis on different convolutional layers separately, following which we merge hierarchical attentional features using a refinement model for better target object representation.

\section{The proposed approach}\label{sec:3}

\subsection{Algorithmic Overview}\label{sec:3.1}

We propose a novel Siamese network model for jointly performing reinforced attentional representation learning and contextual attentional DCF training in an end-to-end fashion. Our network is based on the Siamese network architecture~\cite{siamfc,cfnet}, and takes an image patch pair $(\mathbf{z}, \mathbf{x})$ that comprise a target template patch $\mathbf{z}$ and a searching image patch $\mathbf{x}$ as input. The target template patch $\mathbf{z}$ represents the object of interest that is usually centered at the target object position in the previous video frame. While $\mathbf{x}$ represents the searching region in current video frame, which is centered around the estimated target object position in previous video frame. We use the fully convolutional portion of ResNet-50~\cite{resnet} as the backbone network, and partially modify its original architecture. Both inputs are processed using the same backbone network with learnable parameters  $\varphi$, yielding two deep feature maps, $\varphi(\mathbf{z})$ and $\varphi(\mathbf{x})$. Then, we employ a hierarchical attention module, as proposed in Section~\ref{sec:3.2}, to obtain both the inter- and intra-frame attention of each deep feature hierarchy separately. Subsequently, the hierarchical attentional features are merged using a refinement model. The reinforced attentional representations of $\mathbf{z}$ and $\mathbf{x}$ are denoted as $\varphi^a(\mathbf{z})$ and $\varphi^a(\mathbf{x})$, respectively. The template reinforced attentional representation $\varphi^a(\mathbf{z})$ is used to learn a contextual attentional DCF $\mathbf{w}$ by solving a ridge regression problem $f$ in the Fourier domain~\cite{kcf},
\begin{equation}\label{eq:1}
    \mathbf{w} = f(\varphi^a(\mathbf{z}))
\end{equation}

The contextual attentional DCF $\mathbf{w}$ is then applied to compute the correlation response $g$ of the searching image patch $\mathbf{x}$ as
\begin{equation}\label{eq:2}
    g(\mathbf{x})=\mathbf{w}\star\varphi^a(\mathbf{x})
\end{equation}
where $\star$ denotes the cross-correlation operation. The position of the maximum value of $g$ relative to that of the target object. More details about this part are described in Section~\ref{sec:3.3}.

\subsection{Hierarchical Attention Module}\label{sec:3.2}
\begin{figure}[t]
\begin{center}
\includegraphics[width=\linewidth]{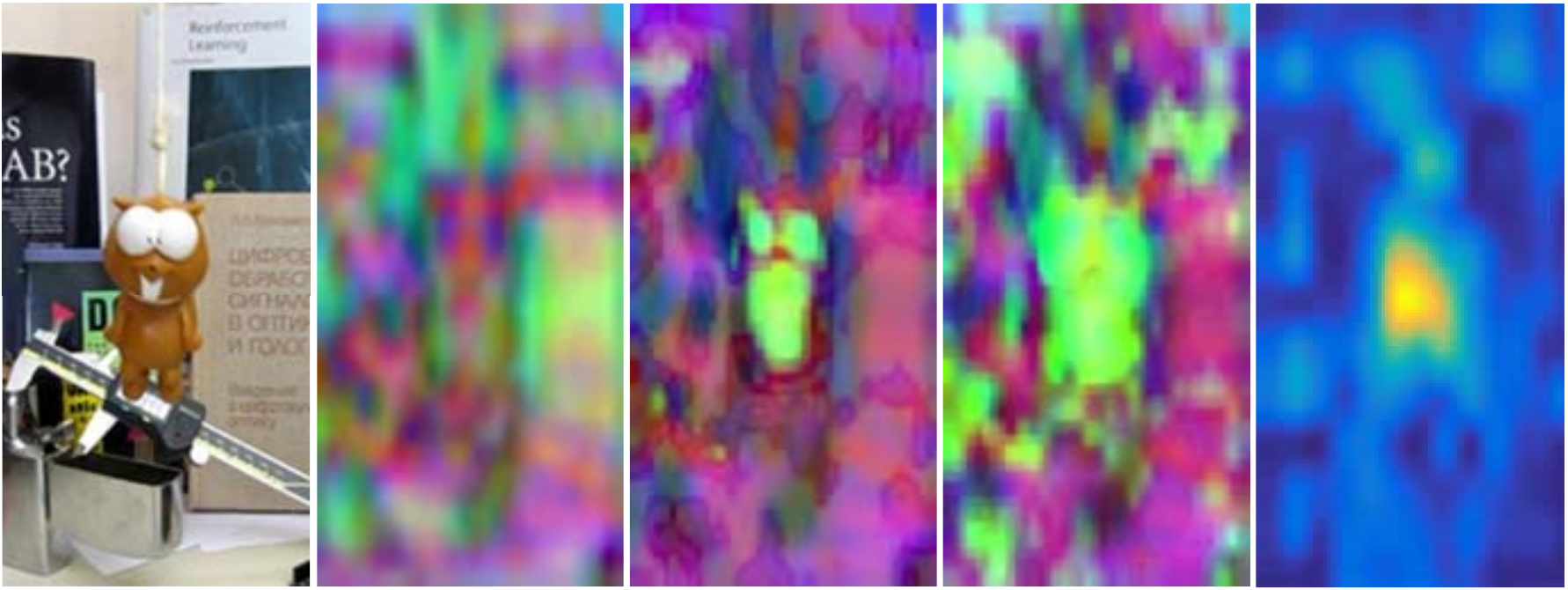}
\end{center}
\caption{Visualization of feature and attention maps of the convolutional layer `\emph{conv}3\_4c' in the backbone network and the correlation response map corresponding to the example image. From left to right are the input example image from the sequence \emph{Lemming}, original feature map, inter-frame attention, intra-frame attention, and the correlation response generated by the proposed network.}
\label{fig:com}
\end{figure}

\begin{figure}
\begin{center}
    \includegraphics[width=\linewidth]{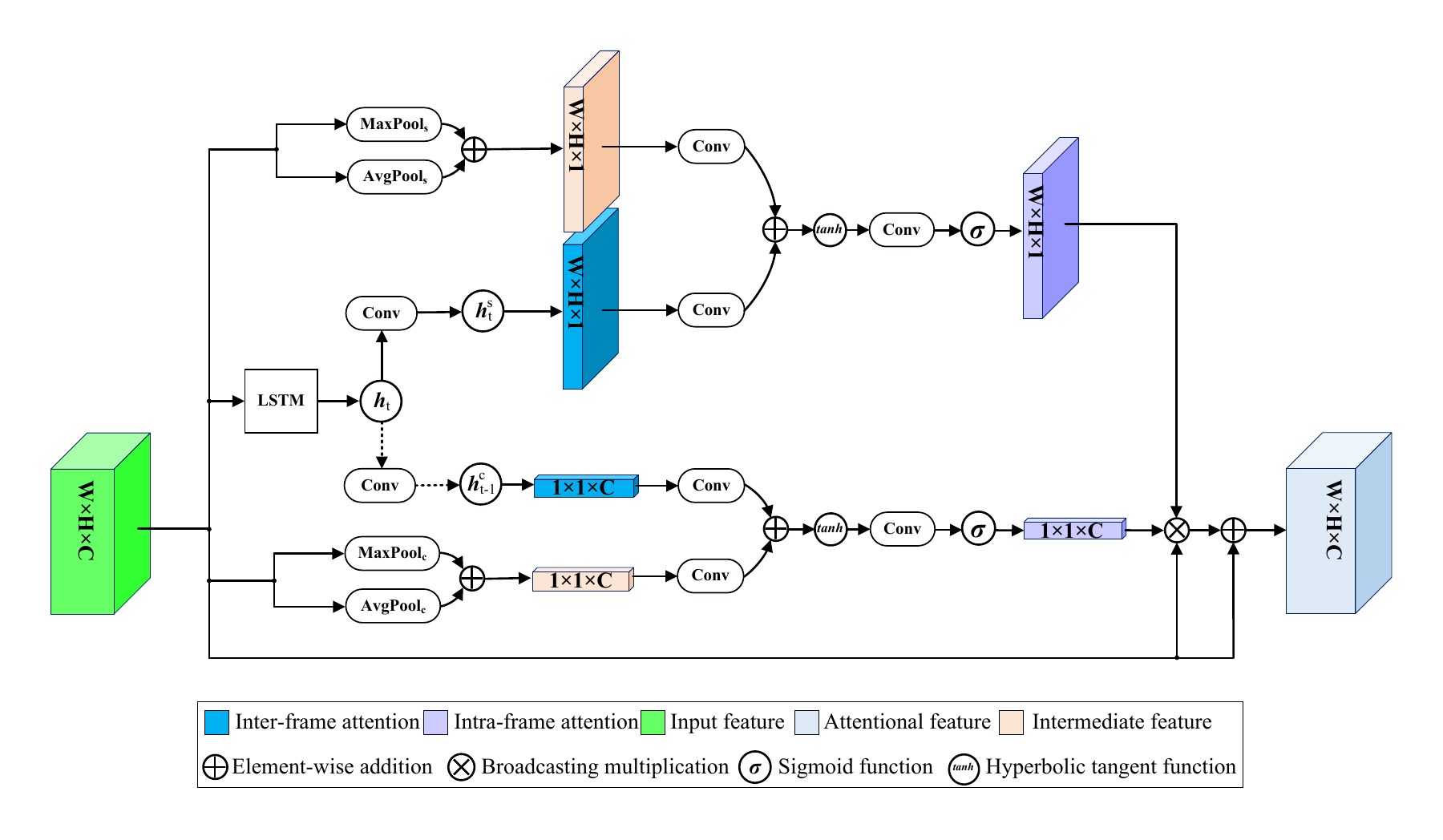}
\end{center}
\caption{Overview of the hierarchical attention module. It is worth noting that $h_{t-1}^c$ is obtained based on the inter-attention $h_{t-1}$ of the previous frame, and we use dashed lines to indicate the corresponding operations.}
\label{fig:intra}
\end{figure}

We introduce a hierarchical attention module to leverage both inter- and intra-frame attention. The inter-frame attention is exploited to perform robust inference in the current frame by capturing historical context information. The intra-frame attention along channel-wise and spatial axes are employed to emphasize the informative representations and suppress  redundancy. As illustrated in Fig.~\ref{fig:com}, for an arbitrary object, the inter-frame attention tends to focus more on some key characteristics of the target object than on the surroundings in consecutive frames (the third picture in Fig.~\ref{fig:com}), while the intra-frame attention mainly concentrates on some critical regions to better represent the target object (the fourth picture in Fig.~\ref{fig:com}). The details of our hierarchical attention module, as shown in Fig.~\ref{fig:intra}, are below.

{\bf Inter-frame attention.} We formulate the tracking task as a sequential inference problem, and utilize a convolutional LSTM unit to model the temporal consistency of the target object’s appearance. On the extracted feature map $\varphi_t\in \mathbb{R}^{W\times H \times C}$ in the current frame $t$, the inter-frame attention can be computed in the convolutional LSTM unit as follows:
\begin{equation}
  \begin{aligned}
    \left( \begin{array}{c}
             f_t \\
             i_t \\
             o_t
           \end{array}\right)&=\sigma(W_{h}h_{t-1}+W_{i}\varphi_t) \\
    \tilde{c}_t &= \tanh(W_{h}h_{t-1}+W_{i}\varphi_t)\\
    c_t &= f_t\odot c_{t-1}+i_t\odot \tilde{c}_t\\
    h_t &= o_t\odot \tanh(c_t)
  \end{aligned}
\end{equation}
where $\oplus$ denotes element-wise addition. $\sigma$ and $\tanh$ are sigmoid activation and hyperbolic tangent activation, respectively. $W_i$ and $W_h$ are the kernel weights of the input layer and the hidden layer. The hyperparameters $f_t$, $i_t$, $o_t$ and $\tilde{c}_t$ indicate the forget, input, output and content gates, respectively. $c_t$ denotes the cell state. $h_t$ is the hidden state that is treated as the inter-frame attention. To facilitate the calculation of the intra-frame attention, $h_t$ is fed into two fully convolutional layers to separately obtain the inter-frame attention along the channel-wise axis $h_t^{c}\in\mathbb{R}^{1\times 1 \times C}$ and the spatial axis $h_t^{s}\in \mathbb{R}^{W\times H \times 1}$,
\begin{equation}
  \begin{aligned}
    h_t^c&=\sigma(W_{hc}h_{t})\\
    h_t^s&=\sigma(W_{hs}h_{t})
  \end{aligned}
\end{equation}
where $W_{hc}$ and $W_{hs}$ are the kernel weights of different convolutional layers corresponding to $h_t^{c}$ and $h_t^{s}$, respectively.

{\bf Intra-frame attention along the channel-wise axis.} We exploit the channel-wise intra-frame attention to make feature maps more visually appealing, and boost the target object discrimination performance. Given the input feature $\varphi_t\in \mathbb{R}^{W\times H \times C}$ and the channel-wise inter-frame attention $h_{t-1}^c$ of the previous frame, we first apply global average-pooling and max-pooling operations along the spatial axis to the input feature to generate two channel-wise context descriptors: $AvgPool_c(\varphi_t)\in \mathbb{R}^{1\times 1 \times C}$ and $MaxPool_c(\varphi_t)\in \mathbb{R}^{1\times 1 \times C}$. Then, we combine and feed them into an MLP with sigmoid activation to obtain the channel-wise intra-frame attention $\Psi_t^c\in \mathbb{R}^{1\times 1 \times C}$ as follows:
\begin{equation}
\begin{aligned}
  \Phi_t^c &= AvgPool_c(\varphi_t)\oplus MaxPool_c(\varphi_t)\\
  \Theta_t^c   &= \tanh\big(W_\Phi^c\Phi_t^c\oplus W_h^ch_{t-1}^c)\\
  \Psi_t^c &= \sigma(W_o^c\Theta_t^c)
\end{aligned}
\end{equation}
where $\sigma$ indicates the sigmoid function, $\oplus$ denotes the element-wise addition. $W_\Phi^c$, $W_h^c$ and $W_o^c$ are weights used to achieve a balance between the dimensions of the channel-wise descriptors and channel-wise intra-frame attention.

{\bf Intra-frame attention along the spatial axis.} We utilize spatial intra-frame attention to highlight target-specific details and enhance the capability for target object localization. Given the input feature $\varphi_t\in \mathbb{R}^{W\times H \times C}$ and the spatial inter-frame attention $h_t^s$ in the current frame, we first combine two different pooled spatial context descriptors $AvgPool_s(\varphi_t)\in \mathbb{R}^{W\times H \times 1}$ and $MaxPool_s(\varphi_t)\in \mathbb{R}^{W\times H \times 1}$. Then, we feed the combination into a MLP using sigmoid activation to generate the spatial intra-frame attention $\Psi_t^s\in \mathbb{R}^{W\times H \times 1}$,
\begin{equation}
\begin{aligned}
  \Phi_t^s &= AvgPool_s(\varphi_t)\oplus MaxPool_s(\varphi_t)\\
  \Theta_t^s   &= \tanh\big(W_\Phi^s\Phi_t^s\oplus W_h^sh_t^s)\\
  \Psi_t^s &= \sigma(W_o^s\Theta_t^s)
\end{aligned}
\end{equation}
where $W_\Phi^s$, $W_h^s$ and $W_o^s$ are the parameters for balancing the dimensions of $\Phi_t^s$ and $\Theta_t^s$. $\sigma$ presents the sigmoid function, and $\oplus$ denotes the element-wise addition.
\begin{figure}
\begin{center}
    \includegraphics[width=.6\linewidth]{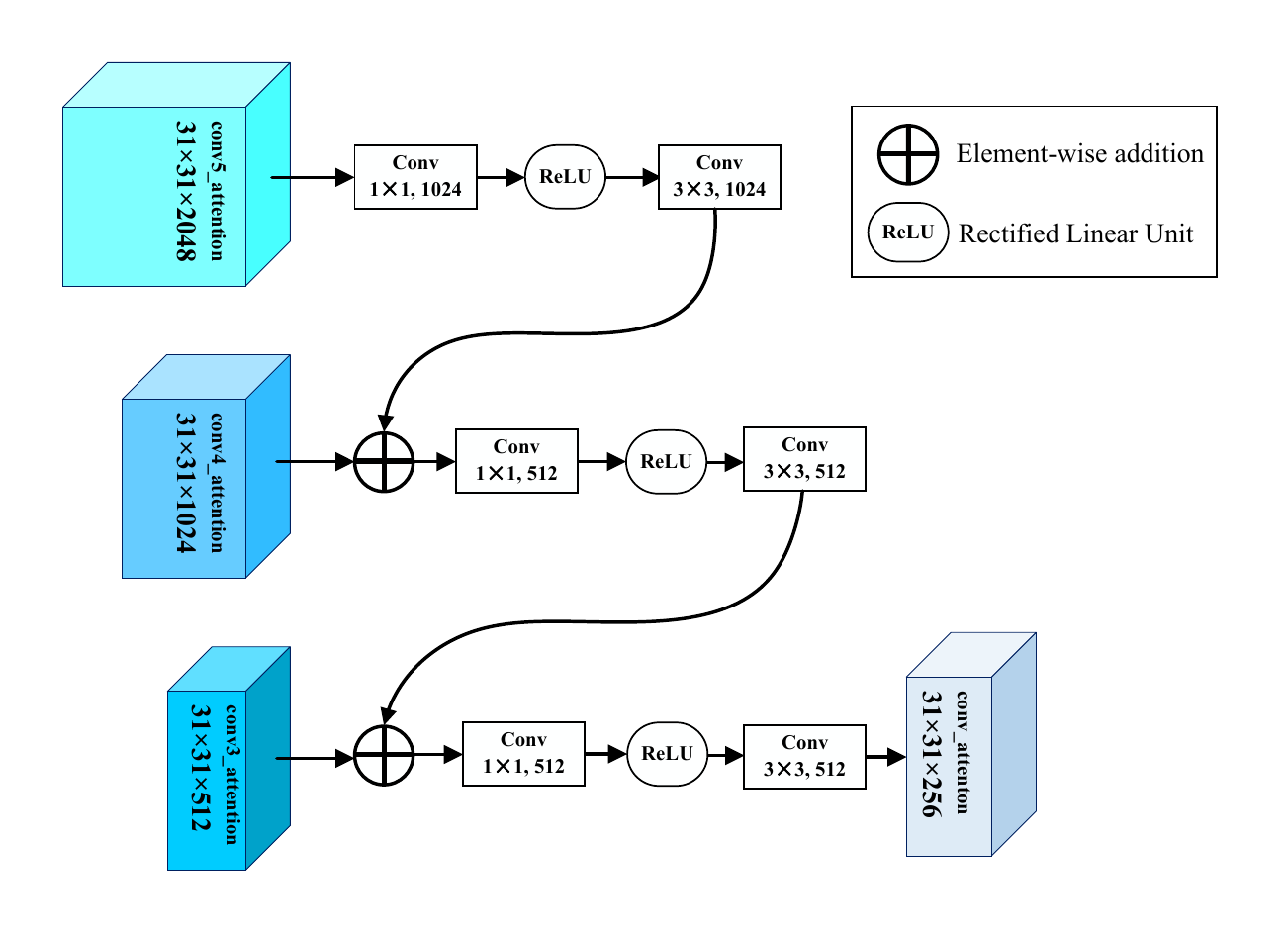}
\end{center}
\caption{Structure of the refinement model.}
\label{fig:refine}
\end{figure}

{\bf Reinforced Attentional Representation} The hierarchical attentional representation $\varphi^a_t$ can be computed using both inter- and intra-frame attention as follows:
\begin{equation}
  \varphi^a_t = \varphi_t \otimes \Psi_t^s \otimes \Psi_t^c \oplus \varphi_t
\end{equation}
where $\oplus$ and $\otimes$ indicate the element-wise addition and broadcasting multiplication, respectively. Finally, we merge those hierarchical attentional representations from coarse to fine to obtain the reinforced attentional representation using a refinement model, as shown in Fig.~\ref{fig:refine}.

\subsection{Contextual Attentional Correlation Layer}\label{sec:3.3}

Unlike traditional DCF-based tracking approaches~\cite{kcf,dsst,csot,eco}, we make some essential modifications to the DCF to utilize the contextual attention in consecutive frames. We choose the context-aware correlation filter (CACF)~\cite{cacf} as the base of our decision module. Because the background around the target object may impact tracking performance, CACF takes the global contextual information into account, and demonstrates outstanding discriminative capability. We crop a target template patch $\mathbf{z}_0$ and $k$ context template patches $\{\mathbf{z}_i\:|\:i=1, 2, \ldots, k\}$ around $\mathbf{z}_0$ from the target template $\mathbf{z}$. Noteworthily, we use a set of target templates from $T$ frames to learn the DCF that has a high response on the target template patch and close to zero for all context template patches,
\begin{equation}\label{eq:4}
    \sum_{t=1}^T\beta_t\Big(\min_\mathbf{w}\|g(\mathbf{z}_{0,t})-\mathbf{y}\|_2^2+\lambda_1\|\mathbf{w}\|_2^2+\lambda_2\sum_{i=1}^{k}\|g(\mathbf{z}_{i,t})\|_2^2\Big)
\end{equation}
where $\beta_t\geq0$ is the impact factor of the target template $\mathbf{z}_t$, $\mathbf{y}$ is the desired correlation response designed as a Gaussian function that centered at the target object position estimated in the previous video frame, $\lambda_2$ controls the context patches regressing to zero. Note that the minimizer in Eq.~\ref{eq:4} is convex, it has a closed-form solution that is given by setting the gradient to zero~\cite{form} as follows:
\begin{equation}\label{eq:3}
    \mathbf{w}=\sum_{t=1}^T\beta_t(\bar{\mathbf{Z}}_{t}^T\bar{\mathbf{Z}}_{t}+\lambda_1\mathbf{I})^{-1}\bar{\mathbf{Z}}_{t}^T\bar{\mathbf{y}}
\end{equation}
where $\bar{\mathbf{Z}}_{t}=[\mathbf{Z}_{0,t},\sqrt{\lambda_2}\mathbf{Z}_{1,t},\sqrt{\lambda_2}\mathbf{Z}_{1,t},\ldots,\sqrt{\lambda_2}\mathbf{Z}_{i,t}]^T$ is a feature matrix. $\mathbf{Z}_{i,t}$ and $\mathbf{Z}_{0,t}$ are circulant feature matrices~\cite{kcf} corresponding to $\mathbf{z}_{i,t}$ and $\mathbf{z}_{0,t}$, respectively. $\bar{\mathbf{y}}=[1,0,0,\ldots,0]$ is the regression objective. For more details, please refer to~\cite{cacf}. The closed-form solution of Eq.~\ref{eq:3} in the Fourier frequency domain can be obtained as follows:
\begin{equation}\label{eq:7}
    \mathbf{\hat{w}}=\frac{\sum_{t=1}^T\beta_t\big(\hat{\varphi}(\mathbf{z}_{0,t})\odot \mathbf{\hat{y}}\big)}
    {\sum_{t=1}^T\beta_t\big(\hat{\varphi}^*(\mathbf{z}_{0,t})\odot\hat{\varphi}(\mathbf{z}_{0,t})+
    \lambda_1+\lambda_2\sum_{i=1}^{k}\hat{\varphi}^*(\mathbf{z}_{i,t})\odot\hat{\varphi}(\mathbf{z}_{i,t})\big)}
\end{equation}
where $\odot$ denotes the Hadamard product, $\mathbf{\hat{w}}$ indicates the discrete Fourier transform $\mathcal{F}(\mathbf{{w}})$, and $\hat{\varphi}^*$ represents the complex conjugate of $\hat{\varphi}$.

Subsequently, the correlation response $g$ in Eq.~\ref{eq:2} can be calculated by performing an exhaustive matching of $\mathbf{w}$ over $\mathbf{x}$ in the Fourier domain as follows:
\begin{equation}\label{eq:5}
g(\mathbf{x})=\mathcal{F}^{-1}\big(\mathbf{\hat{w}}\odot \hat{\varphi}^a(\mathbf{x})\big)
\end{equation}
where $\mathcal{F}^{-1}(\cdot)$ denotes the inverse discrete Fourier transform. Finally, the current target object position can be identified by searching for the maximum value of $g$.

Notably, we formulate the contextual attentional DCF $\mathbf{w}$ as a differentiable correlation layer to achieve the end-to-end training of the whole network and updating the filters online. These capabilities can further enhance the adaptability of our approach to the variations in the target object appearance. Therefore, the network can be trained by minimizing the differences between the real response $g$ and the desired response $\mathbf{y}$ of $\mathbf{x}$. The loss function is formulated as follows:
\begin{equation}
  \mathcal{L}=\|g(\mathbf{x})-\mathbf{y}\|_2^2
\end{equation}

The back-propagation of loss with respect to searching and template patches are computed as follows:
\begin{equation}
  \begin{aligned}
    \frac{\partial\mathcal{L}}{\partial\varphi(\mathbf{x})}     & = \mathcal{F}^{-1}\big( (\hat{g}(\mathbf{x})-\mathbf{\hat{y}})\odot \mathbf{\hat{w}} \big)\\
    \frac{\partial\mathcal{L}}{\partial\varphi(\mathbf{z}_{0})} & = \mathcal{F}^{-1}\big( \frac{\big((\hat{g}(\mathbf{x})-\mathbf{\hat{y}})\odot \mathbf{\hat{z}}_0\big)\odot\big(\mathbf{\hat{y}}-\mathbf{\hat{w}}\odot\hat{\varphi}(\mathbf{z}_{0})\big)}
    {\hat{\varphi}^*(\mathbf{z}_{0})\odot\hat{\varphi}(\mathbf{z}_{0})+
    \lambda_1+\lambda_2\sum_{i=1}^{k}\hat{\varphi}^*(\mathbf{z}_{i})\odot\hat{\varphi}(\mathbf{z}_{i})} \big)\\
    \frac{\partial\mathcal{L}}{\partial\varphi(\mathbf{z}_{i})} & = \mathcal{F}^{-1}\big( \frac{\big((\mathbf{\hat{y}}-\hat{g}(\mathbf{x}))\odot \mathbf{\hat{z}}_i\big)\odot\big(\mathbf{\hat{w}}\odot\hat{\varphi}(\mathbf{z}_{i})\big)}
    {\hat{\varphi}^*(\mathbf{z}_{0})\odot\hat{\varphi}(\mathbf{z}_{0})+
    \lambda_1+\lambda_2\sum_{i=1}^{k}\hat{\varphi}^*(\mathbf{z}_{i})\odot\hat{\varphi}(\mathbf{z}_{i})} \big)
  \end{aligned}
\end{equation}

Once the back-propagation of the correlation layer is derived, our network can be trained end-to-end. The contextual attentional DCF $\mathbf{w}$ is incrementally updated during tracking as formulated in Eq.~\ref{eq:7}.

\section{Experiments}~\label{sec:4}

In this section, we first present the implementation details of our proposed approach. Then, we compare the proposed approach with the state-of-the-art trackers on four modern benchmark datasets, including OTB-2013 with 50 videos~\cite{otb2013}, OTB-2015 with 100 videos~\cite{otb2015}, and VOT-2016~\cite{vot2016} and VOT-2017~\cite{vot2017} both with 60 videos each. Finally, we conduct ablation studies to investigate how the proposed components improve tracking performance.

\subsection{Implementation Details}
\newcommand{\blocka}[2]{\multirow{2}{*}{\(\left[\begin{array}{c}\text{3$\times$3, #1}\\[-.1em] \text{3$\times$3, #1} \end{array}\right]\)$\times$#2}}
\newcommand{\blockb}[3]{\multirow{3}{*}{\(\left[\begin{array}{c}\text{1$\times$1, #2}\\[-.1em] \text{3$\times$3, #2}\\[-.1em] \text{1$\times$1, #1}\end{array}\right]\)$\times$#3}}
\newcommand{\blockc}[3]{\multirow{4}{*}{\text{3$\times$3, maxpool, 2}\\\(\left[\begin{array}{c}\text{1$\times$1, #2}\\[-.1em] \text{3$\times$3, #2}\\[-.1em] \text{1$\times$1, #1}\end{array}\right]\)$\times$#3}}
\begin{table}[t]
\begin{center}
\renewcommand\arraystretch{1}
\caption{Architecture of backbone network. More details of each building blocks are shown in brackets.}
\label{table:backbone}
\begin{tabular}{cccc}
\toprule
\multirow{2}{*}{stage} & output size & \multirow{2}{*}{blocks} & \multirow{2}{*}{stride} \\
  & (input 255$\times$255) & &  \\
\midrule
\multirow{2}{*}{conv1} & \multirow{2}{*}{127$\times$127} & \multirow{2}{*}{7$\times$7, 64}  & \multirow{2}{*}{2} \\
  &  &  &  \\
\hline
\multirow{2}{*}{maxpool1} & \multirow{2}{*}{63$\times$63} & \multirow{2}{*}{3$\times$3} & \multirow{2}{*}{4} \\
  &  &  &  \\
\hline
\multirow{3}{*}{conv2\_x} & \multirow{3}{*}{63$\times$63}  & \blockb{256}{64}{3} & \multirow{3}{*}{4} \\
  &  &  &  \\
  &  &  &  \\
\hline
\multirow{3}{*}{conv3\_x} &  \multirow{3}{*}{31$\times$31}  & \blockb{512}{128}{4}  & \multirow{3}{*}{8}  \\
  &  &  &  \\
  &  &  &  \\
\hline
\multirow{3}{*}{conv4\_x} & \multirow{3}{*}{31$\times$31}  & \blockb{1024}{256}{6}  & \multirow{3}{*}{8}  \\
  &  &  &  \\
  &  &  &  \\
\hline
\multirow{3}{*}{conv5\_x} & \multirow{3}{*}{31$\times$31}  & \blockb{2048}{512}{3}  & \multirow{3}{*}{8}  \\
  &  &  &  \\
  &  &  &  \\
\bottomrule
\end{tabular}
\end{center}
\end{table}

We implement our proposed tracker in Python using MXNet~\cite{mxnet} on an Amazon EC2 instance with an Intel$^\circledR$ Xeon$^\circledR$ E5 CPU @ 2.3GHz with 61GB RAM, and an NVIDIA$^\circledR$ Tesla$^\circledR$ K80 GPU with 12GB VRAM. The average speed of the proposed tracker is $37$ fps. We apply stochastic gradient decent (SGD) with the learning rate varying from $10^{-3}$ to $10^{-4}$, a weight decay of $0.0005$ and a momentum of $0.9$ to train our RAR from scratch on the ImageNet Large Scale Visual Recognition Competition (ILSVRC) video object detection dataset~\cite{imagenet} that has more than $4000$ sequences and $7900$ annotated objects. Table~\ref{table:backbone} illustrates the details of the backbone network, the modified ResNet-50. The deep feature hierarchies extracted from the \emph{conv}3\_4, \emph{conv}4\_6 and \emph{conv}5\_3 block are exploited for visual tracking. To reduce the output feature stride of the original ResNet-50 network from $32$ to $8$, we set the spatial strides to $1$ in the $3\times 3$ convolutional layers of the \emph{conv}4\_1 and \emph{conv}5\_1 block. Thus, all the feature hierarchies have the same spatial resolution. To increase the receptive field, we also adopt deformable convolutions~\cite{defnet} in the $3\times 3$ convolutional layers of the \emph{conv}4\_1 block. Through deformable convolutions~\cite{defnet}, specifically, the deformation of the visual patterns can better fit the target object's structure. The weights of the first two residual stages of the backbone network are fixed, and only the last three residual stages, i.e., \emph{conv}3\_x, \emph{conv}4\_x, and \emph{conv}5\_x, are fine-tuned. The regularization parameters are set as $\lambda_1=10^{-4}$ and $\lambda_2=10^{-1}$. During training, the target template and searching candidates are cropped with a padding size of $2\times$ from two frames picked randomly from the sequence of the same target object, and then resized to a standard input size of $225\times225\times3$. Moreover, to deal with scale variations, we generate a proposal pyramid with three scales $\{a^s\:|\:a=1.02,s\in\big(\lfloor-\frac{S-1}{2}\rfloor,\ldots,\lfloor\frac{S-1}{2}\rfloor\big),S=3\}$ times the previous target object size.

\subsection{Results on OTB}
\begin{table}[t]
\centering
\renewcommand\arraystretch{1}
\caption{Comparisons with recent real-time ($\geq$ 25 fps) state-of-the-art tracking approaches on OTB benchmarks using AUC) and precision metrics. The best three scores are highlighted in \textbf{\textcolor[rgb]{1.00,0.00,0.00}{red}}, \textbf{\textcolor[rgb]{0.00,0.00,1.00}{blue}} and \textbf{\textcolor[rgb]{0.00,1.00,0.00}{green}} fonts, respectively.}
\label{tab:otb}
\resizebox{\textwidth}{!}{%
\begin{tabular}{p{2.8cm}p{1.8cm}<{\centering}p{1.8 cm}<{\centering}p{0.1cm}<{\centering}p{1.8 cm}<{\centering}p{1.8cm}<{\centering}p{0.1cm}<{\centering}p{1.8 cm}<{\centering}}
\toprule
\multirow{2}{*}{Trackers} & \multicolumn{2}{c}{OTB-2013} & & \multicolumn{2}{c}{OTB-2015} && \multicolumn{1}{c}{\multirow{2}{*}{Speed (FPS)}}\\ \cline{2-3}\cline{5-6}
                & AUC & DP && AUC & DP && \\

\midrule
RAR                             & \textbf{\textcolor[rgb]{1.00,0.00,0.00}{0.682}}  & \textbf{\textcolor[rgb]{0.00,0.00,1.00}{0.896}}  && \textbf{\textcolor[rgb]{1.00,0.00,0.00}{0.664}}  & \textbf{\textcolor[rgb]{0.00,0.00,1.00}{0.873}} && 37     \\
DaSiamRPN~\cite{dasiamrpn}      & 0.673  & 0.890  && \textbf{\textcolor[rgb]{0.00,0.00,1.00}{0.658}}  & \textbf{\textcolor[rgb]{1.00,0.00,0.00}{0.881}} && \textbf{\textcolor[rgb]{1.00,0.00,0.00}{97}}      \\
SiamTri~\cite{siamtri}          & 0.615  & 0.815  && 0.590  & 0.781 && \textbf{\textcolor[rgb]{0.00,1.00,0.00}{85}}      \\
SA\_Siam~\cite{sasiam}          & \textbf{\textcolor[rgb]{0.00,0.00,1.00}{0.676}}  & \textbf{\textcolor[rgb]{0.00,1.00,0.00}{0.894}}  && \textbf{\textcolor[rgb]{0.00,1.00,0.00}{0.656}}  & \textbf{\textcolor[rgb]{0.00,1.00,0.00}{0.864}} && 50      \\
SiamRPN~\cite{siamrpn}          & \textbf{\textcolor[rgb]{0.00,1.00,0.00}{0.658}}  & 0.884  && 0.637  & 0.851 && 71      \\
TRACA~\cite{traca}              & 0.652  & \textbf{\textcolor[rgb]{1.00,0.00,0.00}{0.898}}  && 0.602  & 0.816 && 65      \\
EDCF~\cite{edcf}                & 0.665  & 0.885  && 0.635  & 0.836 && 65      \\
CACF~\cite{cacf}                & 0.621  & 0.833  && 0.598  & 0.810 && 33      \\
CFNet~\cite{cfnet}              & 0.611  & 0.807  && 0.568  & 0.767 && 73      \\
SiamFC~\cite{siamfc}            & 0.609  & 0.809  && 0.578  & 0.767 && \textbf{\textcolor[rgb]{0.00,0.00,1.00}{86}}      \\
HCF~\cite{hcft}                 & 0.638  & 0.891  && 0.562  & 0.837 && 26      \\
\bottomrule
\end{tabular}
}
\end{table}

OTB-2013~\cite{otb2013} and OTB-2015~\cite{otb2015} are two popular visual tracking benchmark datasets. The RAR tracker is compared with recent real-time ($\geq$ 25 fps) trackers including DaSiamPRN~\cite{dasiamrpn}, SiamTri~\cite{siamtri}, SA\_Siam~\cite{sasiam}, SiamRPN~\cite{siamrpn}, TRACA~\cite{traca}, EDCF~\cite{edcf}, CACF~\cite{cacf}, CFNet~\cite{cfnet}, SiamFC~\cite{siamfc}, and HCF~\cite{hcft} on these benchmarks. We exploit two evaluation metrics, the distance precision (DP) and overlap success rate (OSR), for comparison. The DP is defined as the percentage of frames where the average Euclidean distance between the estimated target position and the ground-truth is smaller than a preset threshold of 20 pixels, while OSR is the overlap ratios of successful frames exceeded within the threshold range of $[0,1]$. Note that the area under the OSR curve (AUC) is mainly used to assess the different trackers. The evaluation results are illustrated in Table~\ref{tab:otb}.

On the OTB-2013 benchmark dataset, the proposed tracker achieves the best AUC score of $68.2\%$, and the second-best DP score of $89.6\%$. The AUC scores of CACF, CFNet, SiamFC and HCF, the four most related tracking methods to ours, are $62.1\%$, $61.1\%$, $60.9\%$ and $63.8\%$ on the OTB-2013 benchmark dataset, respectively. By comparison, the proposed approach obtains absolute gains of $6.1\%$, $7.1\%$, $7.3\%$ and $4.4\%$. Although the DCF-based tracker TRACA obtains the best DP score of $89.8\%$ on the OTB-2013 benchmark dataset, our RAR tracker outperforms it with an absolute gain of $2.4\%$ in the AUC score. This is because the proposed hierarchical attention strategy used in RAR can best highlight informative representation and suppress redundancy more than the context-aware deep feature compression scheme employed by TRACA. On the OTB-2015 benchmark dataset, our RAR tracker achieves the best AUC score of $66.4\%$ and the second-best DP score of $87.3\%$. However, our tracker does not perform as well as the top-performing DaSiamRPN, which obtains the best DP score of $88.1\%$. This is because DaSiamRPN exploiting extra negative training samples from other datasets to enhance its discriminative capability. We will exploit this training strategy to further boost the performance of RAR. The AUC scores of three other recently published Siamese trackers, SiamTri, SA\_Siam and SiamRPN, on the OTB-2015 benchmark dataset are $59.0\%$, $65.6\%$ and $63.7\%$, respectively. Compared to RAR, their performances drop significantly by more than $7.4\%$, $0.8\%$ and $2.7\%$. This verifies the effectiveness of our network architecture, as the performance of the tracking approach mainly depends on the discriminative capacity of the target object representation. As the baseline of our tracker, EDCF and HCF achieve AUC scores of $63.5\%$ and $56.2\%$ on the OTB-2015 benchmark, respectively. RAR outperforms them by $2.9\%$ and $10.2\%$. In addition, the proposed approach runs efficiently at a real-time speed (37fps).
\begin{figure}
\begin{center}
    \begin{minipage}{\linewidth}\centerline{\includegraphics[width=\linewidth]{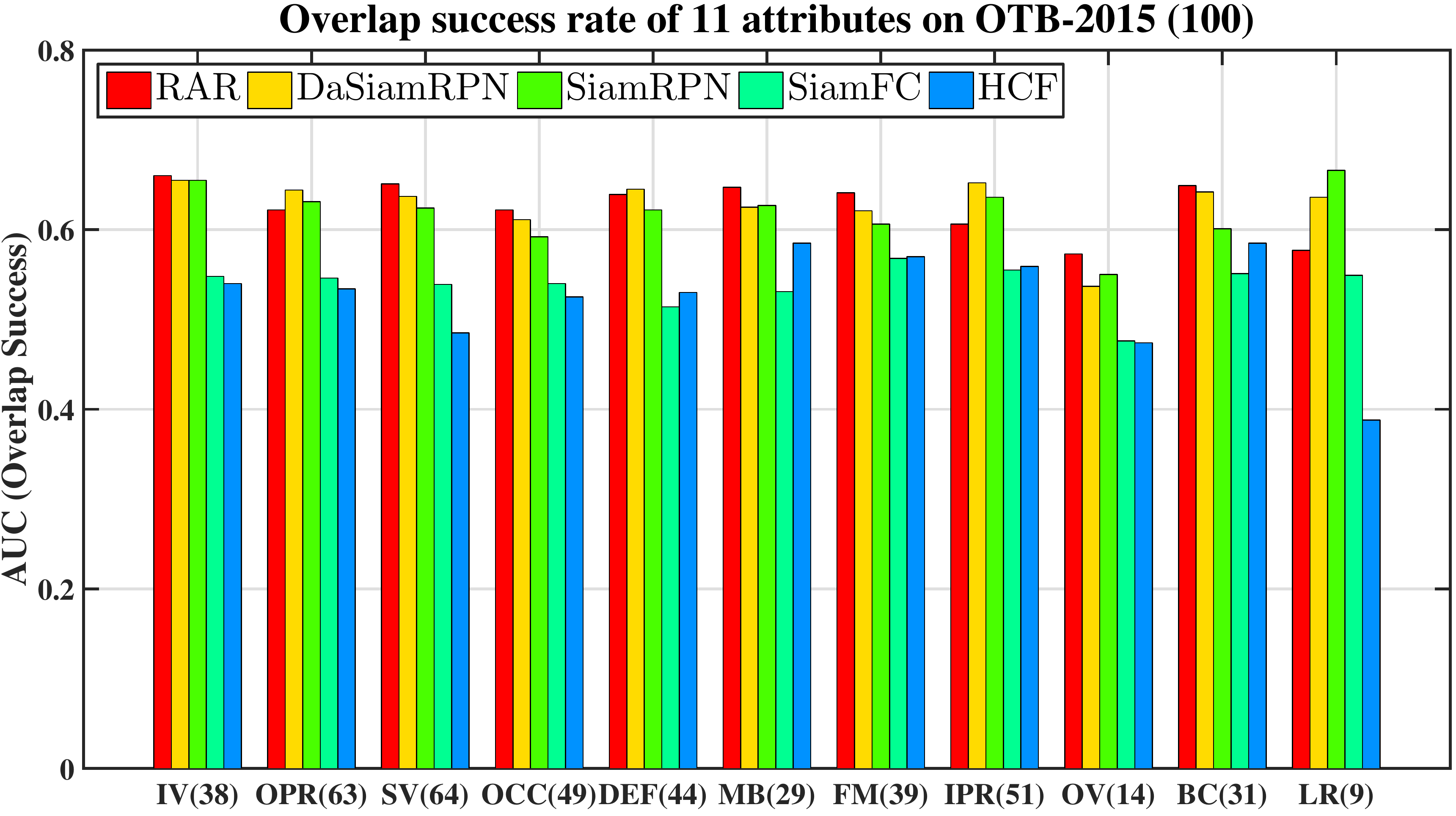}}\end{minipage}
    \vfill\vspace{1em}
    \begin{minipage}{\linewidth}\centerline{\includegraphics[width=\linewidth]{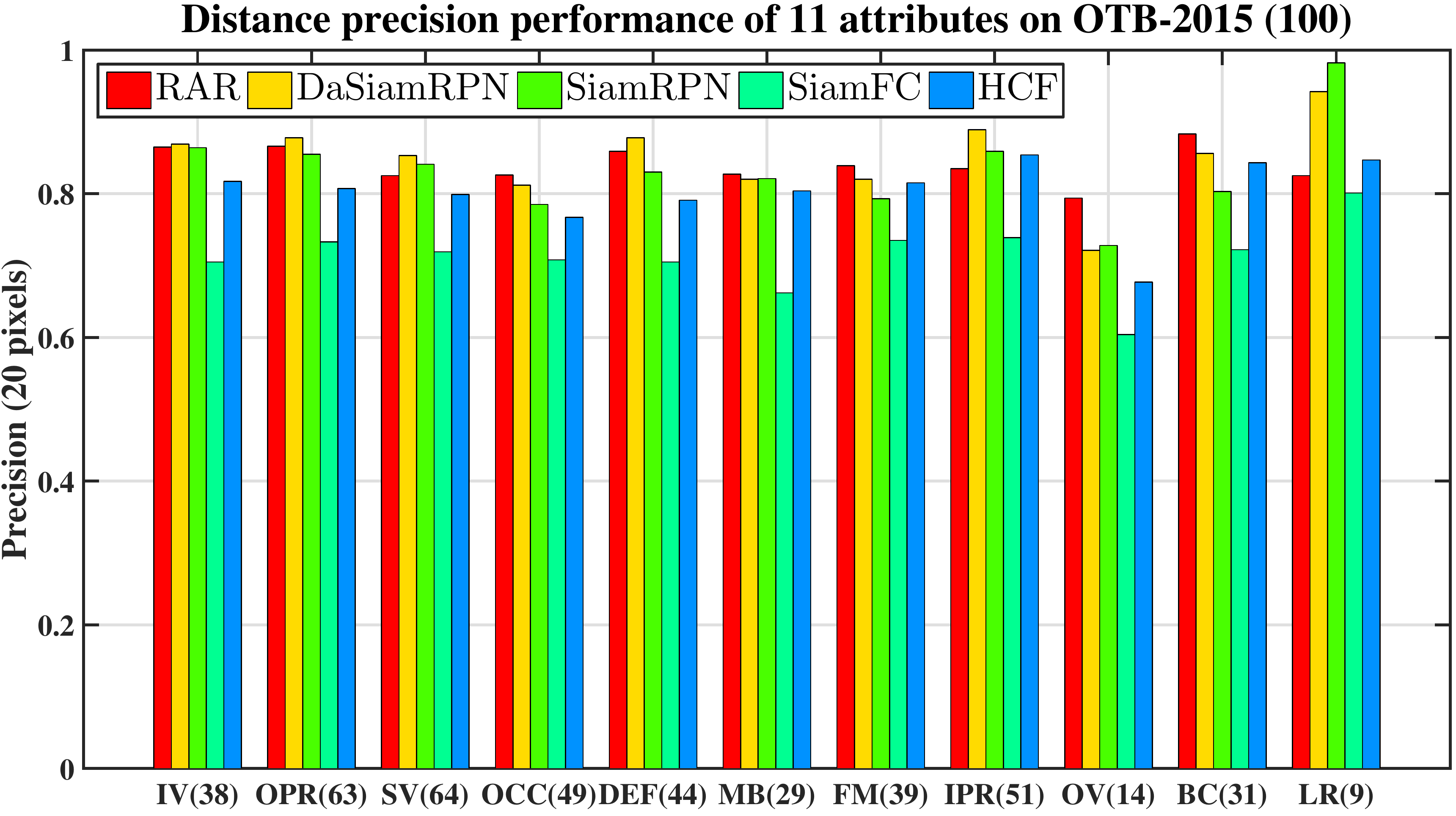}}\end{minipage}
    \vfill
\end{center}
\caption{Performance evaluation of five trackers on the OTB-2015 benchmark dataset with different attributes. Each subset of sequences corresponds to one of the attributes. The later number in the brackets after each attribute acronym is the number of sequences in the corresponding subset.}
\label{fig:attauc}
\end{figure}
\begin{figure}[t]
\begin{center}
    \begin{minipage}{\linewidth}\centerline{\includegraphics[width=\linewidth]{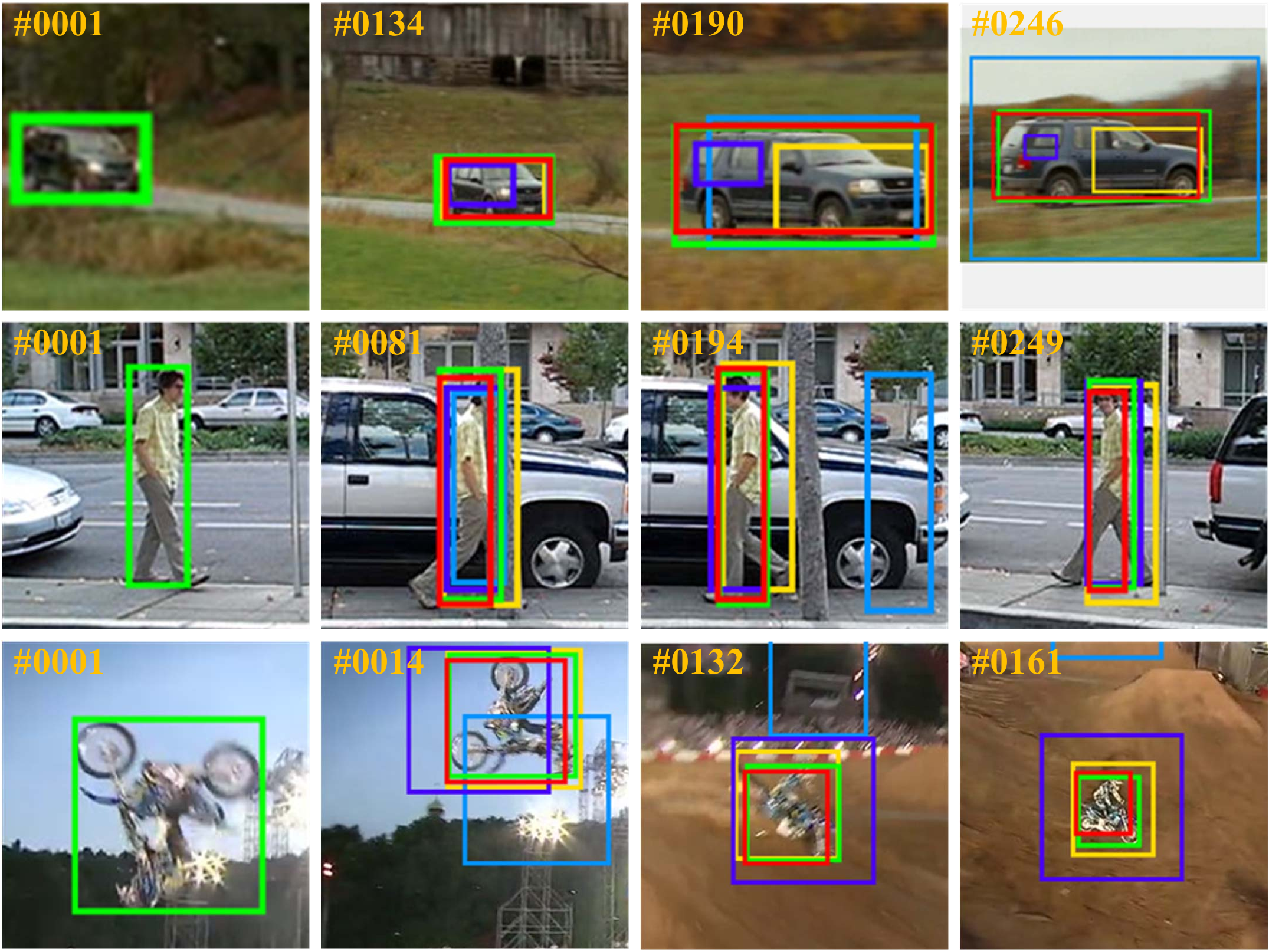}}\end{minipage}
    \vfill\begin{minipage}[b]{0.9\linewidth}\centerline{\includegraphics[width=\textwidth]{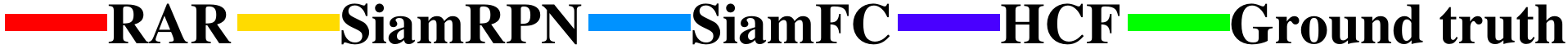}}\end{minipage}
\end{center}
\caption{Comparison of our proposed approach with the state-of-the-art trackers SiamRPN~\cite{siamrpn}, SiamFC~\cite{siamfc}, and HCF~\cite{hcft} on three challenging video sequences (from top to bottom are \emph{carScale}, \emph{david3}, and \emph{MotorRolling}, respectively).}
\label{fig:ress}
\end{figure}

For a comprehensive evaluation, our approach is also compared with state-of-the-art trackers including DaSiamRPN~\cite{dasiamrpn}, SiamRPN~\cite{siamrpn}, SiamFC~\cite{siamfc} and HCT~\cite{hcft} based on different attributes on the OTB-2015 benchmark dataset. The video sequences in OTB are annotated with $11$ different attributes: illumination variation (IV), out-of-plane rotation (OPR), scale variation (SV), occlusion (OCC), deformation (DEF), motion blur (MB), fast motion (FM), in-plane rotation (IPR), out of view (OV), background clutter (BC) and low resolution (LR). The results are presented in terms of AUC and DP scores in Fig.~\ref{fig:attauc}. Although our approach performs worse on three attributes, IPR, OPR, and LR, it achieves impressive performance on the remaining eight attributes. The good performance of the proposed approach can be attributed to two reasons. First, both the inter- and intra-frame attention are effective for selecting more meaningful representation, which accounts for scale and appearance variations. Secondly, with the use of contextual attentional DCF, the proposed approach can further tackle more complicated scenarios such as background clutters and heavy occlusions.

\begin{figure}[t]
\begin{center}
    \begin{minipage}{\linewidth}\centerline{\includegraphics[width=\linewidth]{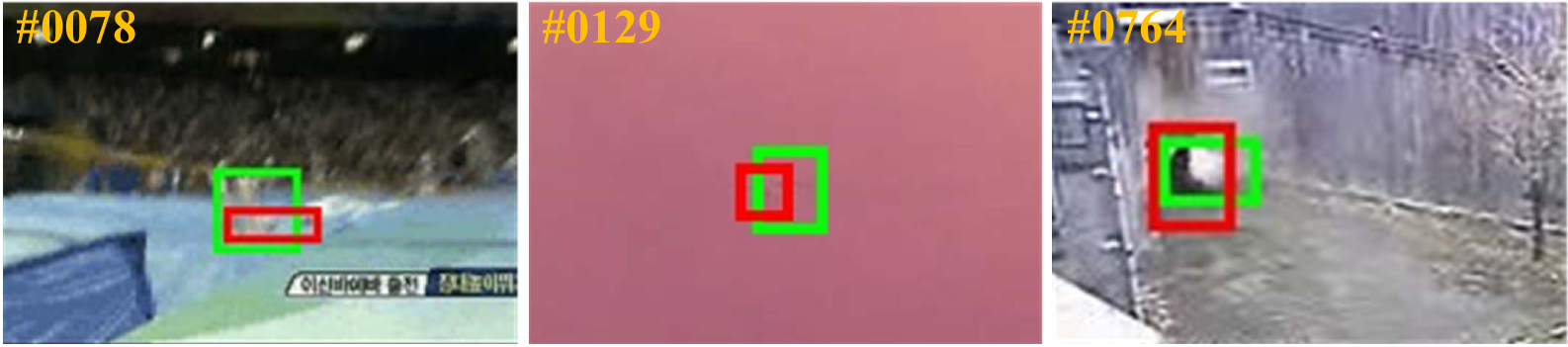}}\end{minipage}
\end{center}
\caption{Failure cases on the \emph{Jump} (IPR occurs at LR), \emph{Bird1} (LR) and \emph{Panda} (OPR occurs at LR) sequences. Red boxes show our results, and the green ones are ground-truths.}
\label{fig:failure}
\end{figure}

Fig.~\ref{fig:ress} shows the comparisons of the proposed approach with excellent trackers SiamRPN~\cite{siamrpn}, SiamFC~\cite{siamfc}, and HCF~\cite{hcft} on three challenging video sequences from the OTB-2015 benchmark dataset. In the sequence \emph{carScale}, the target object undergoes SV with FM. All the trackers, except the proposed one, cannot tackle SV desirably. Both SiamRPN and HCF concentrate on tracking a small part of the target object, while the bounding boxes generated by SiamFC are larger than the ground-truths. In contrast, the proposed approach can trace the target object well. In the sequence \emph{david3}, the target object is partially occluded in a BC scene. SiamFC drifts quickly when OCC occurs, while others are able to trace the target object correctly throughout the sequence. The target object in the sequence \emph{MotorRolling} experiences varying illumination with rotations. Only SiamRPN and RAR can locate the target object accurately. We also reveal three tracking failure cases of our proposed approach in Fig.~\ref{fig:failure}. RAR fails in these sequences mainly because the intra-frame attention cannot learn more meaningful geometries and semantics when IPR/OPR occurs in an LR scene (this is different from the OCC). Thereby, the inter-frame attention plays a dominant role that can still model the temporal coherence within consecutive frames, leading the attentional representation to only focus on some vital parts of the target object.

\subsection{Results on VOT}

\begin{figure}
\begin{center}
    \begin{minipage}{\linewidth}\centerline{\includegraphics[width=\linewidth]{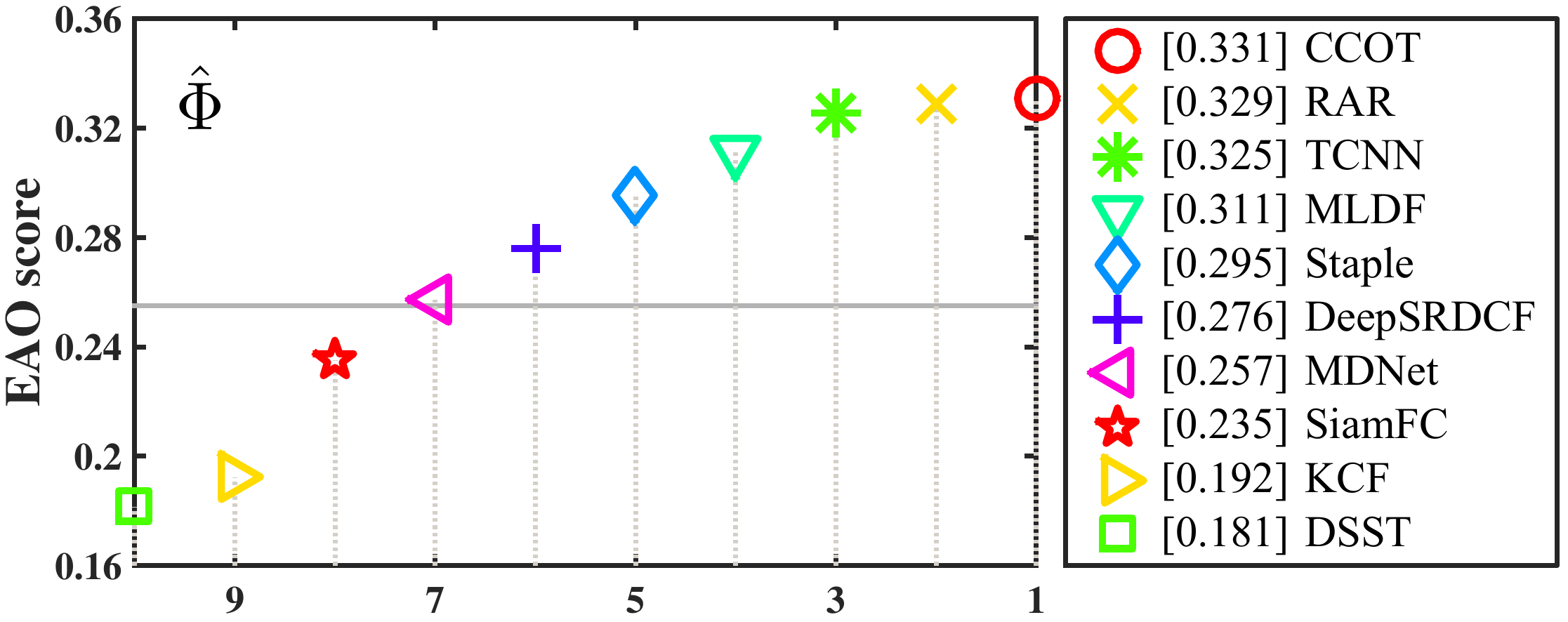}}\centerline{\small (a) EAO scores on VOT-2016}\end{minipage}
    \vfill\vspace{1em}
    \begin{minipage}{\linewidth}\centerline{\includegraphics[width=\linewidth]{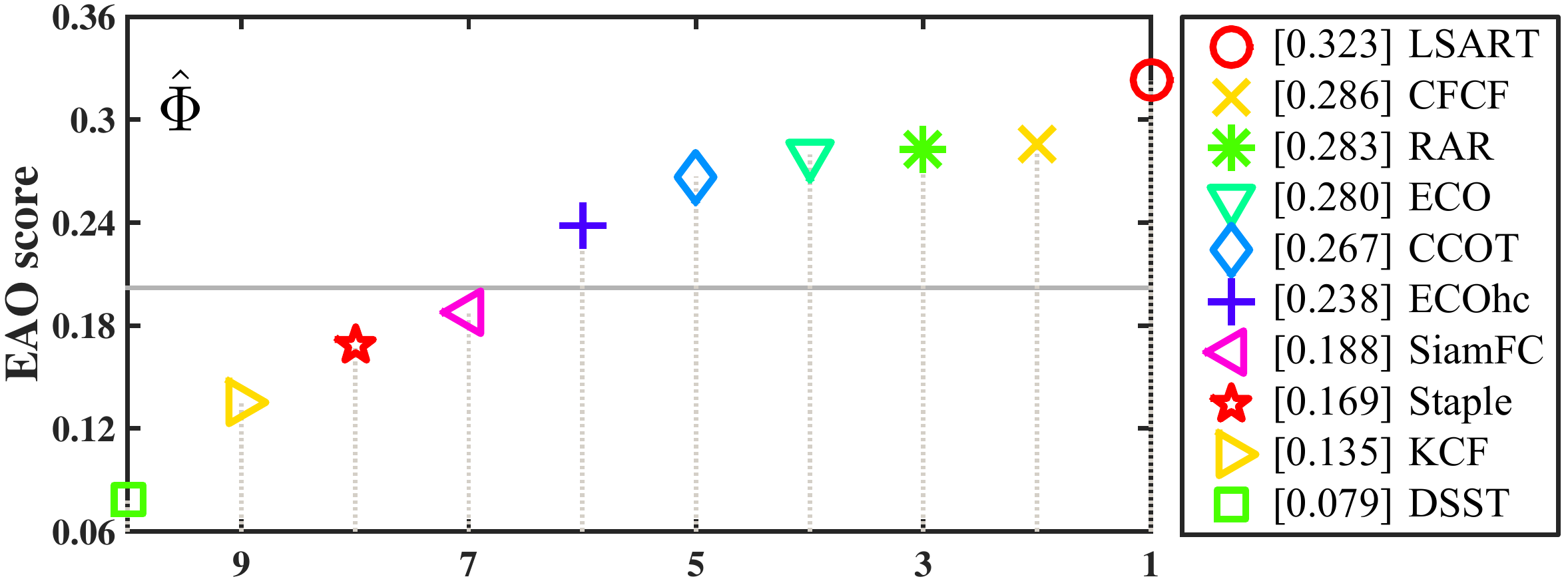}}\centerline{\small (b) EAO scores on VOT-2017}\end{minipage}
\end{center}
\caption{Expected average overlap plot on VOT datasets. The horizontal dashed lines denote the state-of-the-art bounds according to the VOT committee.}
\label{fig:vot}
\end{figure}

The VOT challenge is the largest annual competition in the field of visual tracking. We compare our tracker with several state-of-the-art trackers on the VOT-2016~\cite{vot2016} and VOT-2017~\cite{vot2017} challenge datasets, respectively. Following the evaluation protocol of VOT, we report the tracking performance in terms of expected average overlap (EAO) scores, as shown in Fig.~\ref{fig:vot}.

The RAR tracker obtains the EAO scores of $0.329$ and $0.283$ on these datasets, and outperforms SiamFC~\cite{siamfc} by absolute gains of $9.6\%$ and $9.5\%$, demonstrating its superiority in target object representation. In comparison, CCOT~\cite{ccot} and LSART~\cite{lsart} achieve the top performance on the VOT-2016 and VOT-2017 datasets, respectively. However, LSART runs at $1$ fps, and CCOT runs at approximately $0.3$ fps, our approach runs at orders of magnitude faster than them (37$\times$ and $123\times$). Consequently, our approach exceeds state-of-the-art bounds by large margins, and it can be considered as a state-of-the-art tracker according to the definition of the VOT committee. All the results demonstrate the effectiveness and efficiency of our proposed tracking approach.
\begin{table}[t]
\centering
\renewcommand\arraystretch{1}
\caption{Ablation studies of different configurations of the network backbone (ResNet-50) on the OTB benchmark datasets using AUC and DP scores. C3, C4, and C5 represent \emph{conv}3, \emph{conv}4. and \emph{conv}5 stages, respectively. DefConv indicates whether the convolutional layer exploits deformable convolution. F indicates whether the network backbone is trained offline. S represents the output spatial stride. The best values are highlighted in \textbf{bold} font.}
\label{tab:para}
\resizebox{\textwidth}{!}{%
\begin{tabular}{ccccp{0.01mm}<{\centering}cccccccp{0.01mm}<{\centering}cc}
\toprule
\multirow{2}{*}{\#No.} & \multicolumn{3}{c}{Stage} &  & \multicolumn{3}{c}{DefConv} & \multicolumn{1}{c}{\multirow{2}{*}{F}} & \multirow{2}{*}{S} & \multicolumn{2}{c}{OTB-2013} &  & \multicolumn{2}{c}{OTB-2015} \\ \cline{2-4} \cline{6-8} \cline{11-12} \cline{14-15}
                   &   C3  &   C4  &   C5  & &   C3  &   C4  &   C5  &       &  & AUC & DP  & & AUC & DP  \\\hline
\circled{1}        &$\surd$&       &       & &       &       &       &$\surd$&8 &0.598&0.804& &0.582&0.771\\
\circled{2}        &       &$\surd$&       & &       &       &       &$\surd$&16&0.613&0.807& &0.591&0.778\\
\circled{3}        &       &       &$\surd$& &       &       &       &$\surd$&32&0.581&0.786& &0.557&0.763\\
\circled{4}        &$\surd$&$\surd$&       & &       &       &       &$\surd$&16&0.615&0.825& &0.593&0.791\\
\circled{5}        &$\surd$&       &$\surd$& &       &       &       &$\surd$&32&0.609&0.821& &0.588&0.806\\
\circled{6}        &       &$\surd$&$\surd$& &       &       &       &$\surd$&32&0.626&0.835& &0.603&0.812\\
\circled{7}        &$\surd$&$\surd$&$\surd$& &       &       &       &$\surd$&32&0.634&0.842& &0.617&0.828\\
\circled{8}        &$\surd$&$\surd$&$\surd$& &       &       &       &$\surd$&16&0.653&0.859& &0.629&0.843\\
\circled{9}        &$\surd$&$\surd$&$\surd$& &       &       &       &$\surd$&8 &0.671&0.878& &0.651&0.858\\
\circled{10}       &$\surd$&$\surd$&$\surd$& &       &       &       &$\surd$&4 &0.627&0.816& &0.608&0.794\\
\circled{11}       &$\surd$&$\surd$&$\surd$& &       &       &       &       &8 &0.656&0.843& &0.635&0.822\\
\circled{12}       &$\surd$&$\surd$&$\surd$& &$\surd$&       &       &$\surd$&8 &0.677&0.887& &0.660&0.861\\
\circled{13}       &$\surd$&$\surd$&$\surd$& &       &$\surd$&       &$\surd$&8 &\textbf{0.682}&\textbf{0.896}&&\textbf{0.664}&\textbf{0.873}\\
\circled{14}       &$\surd$&$\surd$&$\surd$& &       &       &$\surd$&$\surd$&8 &0.669&0.872& &0.657&0.860\\
\circled{15}       &$\surd$&$\surd$&$\surd$& &$\surd$&$\surd$&$\surd$&$\surd$&8 &0.675&0.881& &0.653&0.865\\
\bottomrule
\end{tabular}
}
\end{table}
\begin{table}[t]
\centering
\renewcommand\arraystretch{1}
\caption{Ablation studies of several variations of our tracker on the OTB benchmark datasets using AUC and DP scores. The best values are highlighted in \textbf{bold} font.}
\label{table:abla}
\begin{tabular}{p{1.8cm}p{1.1cm}<{\centering}p{1.1cm}<{\centering}p{0.1cm}<{\centering}p{1.1cm}<{\centering}p{1.1cm}<{\centering}}
\toprule
\multirow{2}{*}{Trackers} & \multicolumn{2}{c}{OTB-2013} & & \multicolumn{2}{c}{OTB-2015} \\ \cline{2-3}\cline{5-6}
                & \multicolumn{1}{c}{AUC} & \multicolumn{1}{c}{DP} & & \multicolumn{1}{c}{AUC} & \multicolumn{1}{c}{DP} \\
\midrule
RAR$_{VGG}$     & 0.657  & 0.853  && 0.635  & 0.841      \\
RAR$_{ResNet}$  & 0.634  & 0.842  && 0.617  & 0.828      \\
RAR$_{TDCF}$    & 0.667  & 0.875  && 0.643  & 0.858      \\
RAR$_{NAA}$     & 0.627  & 0.824  && 0.595  & 0.798      \\
RAR$_{NTA}$     & 0.644  & 0.849  && 0.616  & 0.813      \\
RAR$_{NCA}$     & 0.658  & 0.871  && 0.632  & 0.841      \\
RAR$_{NSA}$     & 0.665  & 0.878  && 0.638  & 0.850      \\
\midrule
RAR             & \textbf{0.682}  & \textbf{0.896}  && \textbf{0.664}  & \textbf{0.873}      \\
\bottomrule
\end{tabular}
\end{table}

\subsection{Ablation Studies}

We first modify the backbone network , and conduct an ablation study on the OTB benchmark datasets to reveal the effects of different configurations and parameters of our tracker, including different combinations of feature hierarchies, with or without deformable convolutions and network fine-tuning, and the variations in output feature size (spatial stride). The results are shown in Table~\ref{tab:para}.

We empirically discover that neither the single stage (\circled{1}, \circled{2}, and \circled{3}) nor the combination of two stages (\circled{4}, \circled{5}, and \circled{6}) has achieved competitive performance. After refining the features obtained from all the three stages (\circled{7}), both the AUC and DP scores are steadily improved, with gains of 1.4\% and 1.6\%, compared with the combination of \emph{conv}4 and \emph{conv}5 (\circled{6}) on the OTB-2015 benchmark dataset. This indicates that the feature hierarchies from \emph{conv}3, \emph{conv}4 and \emph{conv}5 have complementary geometries and semantics useful for target object representation. Note that the tracking performance is boosted impressively when the spatial stride is reduced from 32 or 16 to 8 (\circled{9} \emph{vs}. \circled{8} \emph{vs}. \circled{7}). However, as the spatial stride is further reduced from 8 to 4, the performance drops severely (\circled{9} \emph{vs}. \circled{10}). Theoretically, a larger network spatial stride corresponds to a larger receptive field of the neurons in the output stage. A larger receptive field can cover significantly more image context, but is insensitive for target object localization. On the other hand, a smaller receptive field may not be able to capture the target-specific semantics, and inevitably degrades its discriminative capability. This illustrates that the resolution of feature hierarchies is crucial for target object localization and discrimination in visual tracking. We exploit deformable convolution~\cite{defnet} in our approach to adaptively model target object transformations and enlarge the receptive fields. Intriguingly, we observe that merely applying deformable convolution in \emph{conv}4 stages can yield the best performance (\circled{13}), while using the shallow layer (\circled{12}), deeper layer (\circled{14}), and the combination of all the convolutional stages (\circled{15}) achieve only minor performance improvements over the baseline (\circled{9}). That is to say, applying deformable convolution in \emph{conv}3 and \emph{conv}5 stages are not sufficient to enhance the transformation modeling capability and the receptive fields learning adaptability. In addition, our study indicates that fine-tuning the network backbone is necessary (\circled{9} \emph{vs}. \circled{11}), because it yields a great improvement on tracking performance.

To investigate how each proposed component contributes to improving tracking performance, we then evaluate several variations of our approach on the OTB benchmark datasets, including the tracker incorporating the VGG-M network~\cite{vgg} as the backbone (RAR$_{VGG}$); the one deploying the original ResNet-50 network~\cite{resnet} as the backbone (RAR$_{ResNet}$, \circled{7} in the above experiment); the tracker with traditional DCF~\cite{dsst} (RAR$_{TDCF}$); the tracker without hierarchical convolutional features (RAR$_{NHF}$); the tracker not using all the attention (RAR$_{NAA}$); and the trackers not deploying any single attention (RAR$_{NTA}$ means we do not use the inter-frame attention; RAR$_{NCA}$ means we do not use the channel-wise inter- and intra-frame attention; and RAR$_{NSA}$ means we do not use the spatial inter- and intra-frame attention). The detailed evaluation results are illustrated in Table~\ref{table:abla}.

Our full algorithm (RAR) outperforms all those variants. RAR achieves absolute gains of $2.9\%$ and $4.7\%$ in the AUC scores, compared with RAR$_{VGG}$ and RAR$_{ResNet}$ on the OTB-2015 benchmark dataset, respectively. Therefore, it has been proven that our modified backbone network learns more informative target object representation by enhancing the generalization capability. It is worth noting that RAR$_{ResNet}$ underperforms RAR$_{VGG}$ with $1.9\%$ drop. This performance degradation can be directly attributed to both the receptive field and the output stride of ResNet-50 are too large to capture more useful information, even though the architecture of ResNet-50 is deeper than VGG-M. To evaluate the impact of the hierarchical attention mechanism, we remove it, and directly use the original deep features from three stages to represent the target objects. This elimination causes remarkable performance drops, i.e., a degradation of $6.9\%$ in the AUC score from 0.664 to 0.595 on the OTB-2015 benchmark dataset. It clearly confirms the effectiveness of the combination of inter- and intra-frame attention for emphasizing meaningful representations and suppressing redundant information. Besides, by introducing the differentiable correlation layer, the AUC score can be significantly increased by $1.5\%$ compared with RAR$_{TDCF}$ on the OTB-2013 benchmark dataset. This performance gain demonstrates the superiority of the proposed contextual attentional DCF. According to our ablation studies, every component in our approach contributes to improving tracking performance.

\section{Conclusions}\label{sec:5}

In this paper, we propose an end-to-end network model that can jointly achieve hierarchical attentional representation learning and contextual attentional DCF training for high-performance visual tracking. Specifically, we introduce a hierarchical attention module to learn hierarchical attentional representation using both inter- and intra-frame attention at different convolutional layers to emphasize informative representations and suppress redundant information. Moreover, a contextual attentional correlation layer is incorporated into the network to enhance the tracking performance for accurate target object discrimination and localization. Experimental results clearly demonstrate that our proposed tracker significantly outperforms most state-of-the-art trackers both in terms of accuracy and robustness at a speed above the real-time requirement. Although the proposed tracker has achieved competitive tracking performance, it can be further improved by utilizing multimodal representation and robust backbone networks, such as natural linguistic features or graph convolutional networks.

\end{spacing}

\bibliography{arxiv}

\end{document}